\pdfoutput=1

\documentclass[11pt]{article}

\usepackage{acl}

\usepackage{times}
\usepackage{latexsym}

\usepackage[T1]{fontenc}

\usepackage[utf8]{inputenc}

\usepackage{microtype}

\usepackage{amsmath,amsfonts,bm,amssymb}
\usepackage{booktabs} %
\usepackage{url}
\usepackage{graphicx}
\usepackage{array}
\usepackage{color}
\usepackage{makecell}
\usepackage{multirow}
\usepackage{xspace}
\usepackage{colortbl}
\usepackage{arydshln}
\usepackage{pifont}
\usepackage[export]{adjustbox}

\newcommand{\ie}{\textit{i.e.}} %
\newcommand{\eg}{\textit{e.g.}} %
\newcommand{\start}[1]{\vspace{.0mm}\noindent{{\bf #1}.}}

\newcommand{\upv}{\vspace{-.0cm}}
\newcommand{\downv}{\vspace{-.1cm}}

\newcommand{\Catts}{\textsc{Catts}\xspace}
\newcommand{\Cattsxsum}{\textsc{Catts}$_\mathrm{XSUM}$\xspace}
\newcommand{\bart}{\textsc{BART}\xspace}
\newcommand{\bartxsum}{\textsc{BART}$_\mathrm{XSUM}$\xspace}

\newcommand{\ours}{\textsc{CiTeS}\xspace}
\newcommand{\oursTitle}{$\textsc{CiTeS}_{\text{Title}}$\xspace}
\newcommand{\ourdata}{CiteSum\xspace}

\definecolor{gred}{RGB}{219,68,55}
\definecolor{gblue}{RGB}{66,133,244}
\definecolor{gyellow}{RGB}{244,180,0}
\definecolor{ggreen}{RGB}{15,157,88}
\definecolor{ggrey}{RGB}{115,115,115}
\definecolor{na}{gray}{0.9}

\newcommand{\colorR}[1]{\textcolor{gred}{\textbf{#1}}}
\newcommand{\colorG}[1]{\textcolor{ggreen}{\textbf{#1}}}
\newcommand{\colorB}[1]{\textcolor{gblue}{\textbf{#1}}}

\title{\ourdata: Citation Text-guided Scientific Extreme Summarization\\ and Domain Adaptation with Limited Supervision}

\author{\makecell{Yuning Mao, Ming Zhong, Jiawei Han} \\
University of Illinois Urbana-Champaign  \\ \{yuningm2, mingz5, hanj\}@illinois.edu 
} 

\begin{document}
\maketitle
\begin{abstract}
Scientific extreme summarization (TLDR) aims to form ultra-short summaries of scientific papers.
Previous efforts on curating scientific TLDR datasets failed to scale up due to the heavy human annotation and domain expertise required.
In this paper, we propose a simple yet effective approach to automatically extracting TLDR summaries for scientific papers from their citation texts.
Based on the proposed approach, we create a new benchmark \ourdata without human annotation, which is around 30 times larger than the previous human-curated dataset SciTLDR.
We conduct a comprehensive analysis of \ourdata, examining its data characteristics and establishing strong baselines.
We further demonstrate the usefulness of \ourdata by adapting models pre-trained on \ourdata (named \ours) to new tasks and domains with limited supervision.
For scientific extreme summarization, \ours outperforms most fully-supervised methods on SciTLDR without any fine-tuning and obtains state-of-the-art results with only 128 examples.
For news extreme summarization, \ours achieves significant gains on XSum over its base model (not pre-trained on \ourdata), \eg, +7.2 ROUGE-1 zero-shot performance and state-of-the-art few-shot performance.
For news headline generation, \ours performs the best among unsupervised and zero-shot methods on Gigaword.\footnote{Our dataset and code can be found at \url{https://github.com/morningmoni/CiteSum}.}
\end{abstract}

\section{Introduction}
Scientific summarization typically regards paper abstract as the ground-truth summary, as it is written by the authors themselves with relatively high quality and readily available in most scientific documents.
However, paper abstract may not always be the ideal summary because it often involves certain details such as task description, background information, and experiment results (cf. the abstract of this paper).
As a result, recent work \cite{cachola-etal-2020-tldr} has studied the problem of scientific extreme summarization, which aims at forming ultra-short summaries (usually one sentence) of the papers, namely the TLDR\footnote{``TLDR'' (or ``TL;DR'') is short for ``too long; didn't read'', often used in online discussions about scientific papers.} summaries.

\begin{table}[t]
        \resizebox{1.0\columnwidth}{!}{
        \begin{tabular}{p{8.5cm}}
            \toprule
                \textbf{Paper Abstract}: We study the problem of \colorB{transferring a sample in one domain to an analog sample in another domain}. Given two related domains, S and T , we would like to learn a generative function G that maps an input sample from S to the domain T , such that the output of a given function f , which accepts inputs in either domains, would remain unchanged. Other than the function f , the training data is unsupervised and consist of a set of samples from each domain. \colorG{The Domain Transfer Network (DTN)} we present employs a compound loss function that includes a multiclass GAN loss, an f -constancy component, and a regularizing component that encourages G to map samples from T to themselves. We apply our method to visual domains including \colorR{digits and face images} and demonstrate its ability to generate convincing novel images of previously unseen entities, while preserving their identity. \\
            \midrule
                \textbf{Citation Text}: Taigman et al. [8] proposed \colorG{the Domain Transfer Network (DTN)} to \colorB{map a sample from one domain to an analog sample in another domain} and achieved favorable performance on small resolution \colorR{face and digit images}. \\
 
            \bottomrule
        \end{tabular}
        }
    \upv
    \caption{An example showing that the citation texts of a paper can often be used as its ultra-short summary. }
    \label{table_example}
    \downv
    \end{table}

However, unlike paper abstracts, ultra-short paper summaries are far from being universally available. 
Only certain scientific venues such as OpenReview.net support a \textit{TLDR} field during paper submission, which is completely optional, and not all submitted papers provide such information.
In addition, human-annotated summaries of scientific documents are rather costly and require domain expertise.
As a consequence, the previous SciTLDR dataset \cite{cachola-etal-2020-tldr}, using a combination of author-provided TLDR and human-annotated TLDR (rephrased from paper reviews on OpenReview), only collected around 2,000 examples for training and 600 for testing.

In this paper, we argue that citation texts can often serve as high-quality short summaries of the cited papers.
In Table~\ref{table_example}, we show the abstract of one paper and its citation sentence in a follow-up paper.
We observe that the citation sentence introduces the cited method and its contributions in a concise and accurate manner.
Motivated by such observations, we propose a simple yet effective approach to locating, extracting, and filtering citation texts from scientific papers.
We then treat the processed citation texts as ground-truth summaries of the cited papers.
Based on the proposed approach, we create a large-scale scientific extreme summarization benchmark, \ourdata,  which is automatically derived from citation texts and around 30 times larger than the previous human-annotated dataset SciTLDR \cite{cachola-etal-2020-tldr}.

We conduct a comprehensive analysis of \ourdata regarding its data characteristics and quality, meanwhile establishing strong baselines as the reference for future studies.
We further verify the usefulness of \ourdata by demonstrating that models pre-trained on \ourdata, which we name as \ours (\textbf{Ci}tation \textbf{Te}xt-guided \textbf{S}ummarizer), exhibit superior generalizability during low-resource adaptation to new tasks and domains.

On the human-annotated scientific extreme summarization dataset SciTLDR \cite{cachola-etal-2020-tldr}, our zero-shot BART-based \cite{lewis-etal-2020-bart} \ours, without any fine-tuning, performs better than most \textit{fully-supervised} baselines, including the fully-supervised BART model (without pre-training on \ourdata).
Our few-shot \ours achieves state-of-the-art performance with only 128 labeled examples from SciTLDR.
In addition, \ours outperforms its base model (BART) on two more diverse scientific tasks -- discipline classification and title generation.
When transferring to news extreme summarization, despite the domain mismatch, \ours achieves significantly better zero-shot performance than BART and PEGASUS \cite{zhang2020pegasus} (\eg, +7.2 ROUGE-1) and state-of-the-art few-shot performance on the XSum dataset \cite{narayan-etal-2018-dont}.
Furthermore, \ours performs the best among unsupervised and zero-shot methods on the Gigaword news headline generation dataset  \cite{rush-etal-2015-neural}.

\start{Contributions}
(1) We propose a simple yet effective approach to automatically extracting ultra-short paper summaries from citation texts.
(2) Based on the proposed approach, we create a large-scale scientific extreme summarization benchmark \ourdata and conduct a comprehensive analysis of it.
(3) We further verify the quality and usefulness of \ourdata by demonstrating that models pre-trained on \ourdata perform very well on new tasks and domains such as news extreme summarization and headline generation with limited training.

\section{\ourdata: A Large-scale Scientific Extreme Summarization Benchmark}

\subsection{Data Creation}

\paragraph{Data Source}
We take the publicly available Semantic Scholar Open Research Corpus (S2ORC) \cite{lo-etal-2020-s2orc} as the source for data creation.
In the latest version of S2ORC, there are 136M scientific papers from different academic disciplines and the number of papers with full-text access is 12M.
We further remove papers without citation information, resulting in 9M papers as the candidates.

\begin{table}[t]
        \resizebox{1.0\columnwidth}{!}{
        \begin{tabular}{p{8cm}}
            \toprule
                \textbf{Citation Example 1}: We take the publicly available Semantic Scholar Open Research Corpus (S2ORC) \cite{lo-etal-2020-s2orc} as the source for data creation. \\
            \midrule
                \textbf{Citation Example 2}: Unlike WikiTransfer \cite{fabbri-etal-2021-improving}, \ours does not involve any downstream task-specific data selection or model tuning.\\
            \bottomrule
        \end{tabular}
        }
    \upv
    \caption{Examples (in our paper) showing that citation texts have different intents and cannot \textit{always} be used as summaries of the cited paper. }
    \label{table_bad_example}
    \downv
    \end{table}

\begin{table*}[t]
\vspace{-0.2cm}
\centering

\resizebox{1.8\columnwidth}{!}{
\scalebox{1}{
\begin{tabular}{lrrr c}
\toprule
\textbf{Dataset} & \textbf{Train / Val / Test} & \textbf{len$_{\text{src}}$}  & \textbf{len$_{\text{summ}}$} & \textbf{Automatic?} \\
\midrule
Gigaword \cite{rush-etal-2015-neural} & 3,803,957	/ 189,651 / \ 1,951 & 32 & 9 &\ding{51}\\
XSum \cite{narayan-etal-2018-dont} & 204,045 / 11,332 / 11,334 & 431 & 23 &\ding{51}\\
arXiv \cite{cohan-etal-2018-discourse}  &   203,037 / \ \ 6,436 /\ \ \ 6,440  &   4.9K & 220 & \ding{51}\\
SciSummNet \cite{yasunaga2019scisummnet} & \ \ \ \ 1,000 /\ \ \quad - \quad / \ \quad- \quad \quad  & 4.7K & 150 & \ding{55}\\
TalkSumm \cite{lev-etal-2019-talksumm} & \ \ \ \ 1,716 /\ \ \quad - \quad / \ \quad- \quad \quad  & 4.8K & 150 & \ding{51} \\ 
SciTLDR \cite{cachola-etal-2020-tldr} & \ \ \ \ 1,992 /\ \ \ \ \ \ 619 / \ \ \ \ \ 618	& 159  & 21 & \ding{55}\\
\cellcolor{na}\ourdata & \ \ 83,304 / \ \  4,721 / \ \ 4,921 &  255 & 23 & \ding{51}\\

\bottomrule
\end{tabular}
}
}
\upv
\caption{Statistics of relevant summarization datasets showing the number of samples per data split, the average number of words in the source document (src) and reference summary (summ), and whether dataset creation is automatic without human annotation.
SciSummNet \cite{yasunaga2019scisummnet} and TalkSumm \cite{lev-etal-2019-talksumm} do not contain validation/test set as their model evaluation was done on another dataset \cite{jaidka-etal-2016-overview}.
}
\label{tab:dataset}
\downv
\end{table*}

\paragraph{Quality Control}
Not all citation texts are of high quality and can be used as summaries of the cited papers.
In Table~\ref{table_bad_example}, we show two examples (in our paper) where the citation sentence simply (1) describes the data source or (2) introduces the difference of the citing paper from the cited paper.
We note that prior studies on citation text generation \cite{chen-etal-2021-capturing,ge-etal-2021-baco} often do not filter these citation texts and simply treat all paragraphs/sentences with citations as the ground-truth labels, as their goals are not on paper summarization but writing assistance.

To ensure data quality, we carefully locate, extract, and filter the citation texts of papers in the following manner.
First, we only take citation texts in the Related Work section of a paper, which largely ensures that they describe the content of the cited paper instead of irrelevant information, such as task background in the Introduction section or the implementation details in the Experiment section.
After filtering papers without a Related Work section, there are around 288K papers left.

Second, we only keep citation sentences that cite a single paper, since those with multiple citations typically discuss one line of work and cannot be used as the summary of a specific paper.
In total, we obtain about 426K citation sentences.

Next, we measure the similarity between the citation texts and the cited papers and filter dissimilar pairs.
Intuitively, if a citation sentence can serve as a high-quality summary, certain amount of its content should be from the cited paper.
Prior work~\cite{lu-etal-2020-multi-xscience} also showed that authors tend to cite a paper using the information in the abstract of the cited paper.
We thus calculate the overlap between paper abstracts and their citation sentences, and filter those below a threshold $\mathcal{T}$.
We set $\mathcal{T}$ to 50/20/40 for ROUGE-1/2/L recall through manual examination, resulting in a ROUGE-1/2/L recall of 73.1/39.4/58.5 after filtering.\footnote{We also experimented with semantic metrics such as BERTScore \cite{zhang2019bertscore} but they did not function as well as ROUGE-based metrics in our human evaluation.}
As a reference, the ROUGE-1/2/L recall between paper abstracts and reference summaries on SciTLDR \cite{cachola-etal-2020-tldr} is 81.1/38.9/62.0 and 65.2/17.9/45.7 for author-provided (SciTLDR-Auth) and peer review-derived (SciTLDR-PR) TLDR, respectively.
That is, the abstraction level of \ourdata is between SciTLDR-Auth and SciTLDR-PR.
This filtering step is rather strict as we prefer quality to quantity of the data and only 93K of the 426K examples (21.8\%) are kept.

We further replace each citation span (\eg, ``Taigman et al. [8]'') with a special token ``REF'' as they vary in different papers but essentially have the same meaning (\ie, referring to a cited paper).

\paragraph{Dataset Split}
After data filtering and preprocessing,  there are 92,946 examples in the final \textit{citation text-guided summarization} dataset, which we name as \ourdata.
We take about 5\% of the data as the validation and test sets respectively, and the remaining 90\% as the training set.
As one paper may be cited multiple times in different papers, we ensure that there is no label leakage by excluding papers used for evaluation from the training set.

\subsection{Data Analysis}

\paragraph{Dataset Statistics}

In Table~\ref{tab:dataset}, we show the data statistics of \ourdata and other relevant summarization datasets.
In terms of data size, \ourdata is about half the size of other automatically constructed datasets like XSum \cite{narayan-etal-2018-dont} and arXiv \cite{cohan-etal-2018-discourse} due to the availability of citation texts and our strict quality control.
On the other hand, the size of \ourdata is much larger than \textit{human-annotated} datasets on paper summarization \cite{yasunaga2019scisummnet,cachola-etal-2020-tldr} -- almost 30 times larger than the SciTLDR dataset \cite{cachola-etal-2020-tldr}.

When compared to SciTLDR, the average length of source documents in \ourdata is longer, while that of the reference summaries is similar as the majority of summaries in SciTLDR also involve one sentence.
When compared to XSum, the summary length in \ourdata is also quite similar.
However, the inputs in XSum are news articles instead of scientific papers and the input lengths also vary.
As for Gigaword \cite{rush-etal-2015-neural}, a news headline generation dataset, both its source input and target output are much shorter than \ourdata.
Despite such differences, we observe that our models pre-trained on \ourdata transfer very well to these datasets in zero-shot and few-shot settings (Sec.~\ref{sec_exp_transfer}).

\begin{figure}[t]
    \includegraphics[width=1.06\linewidth,center]{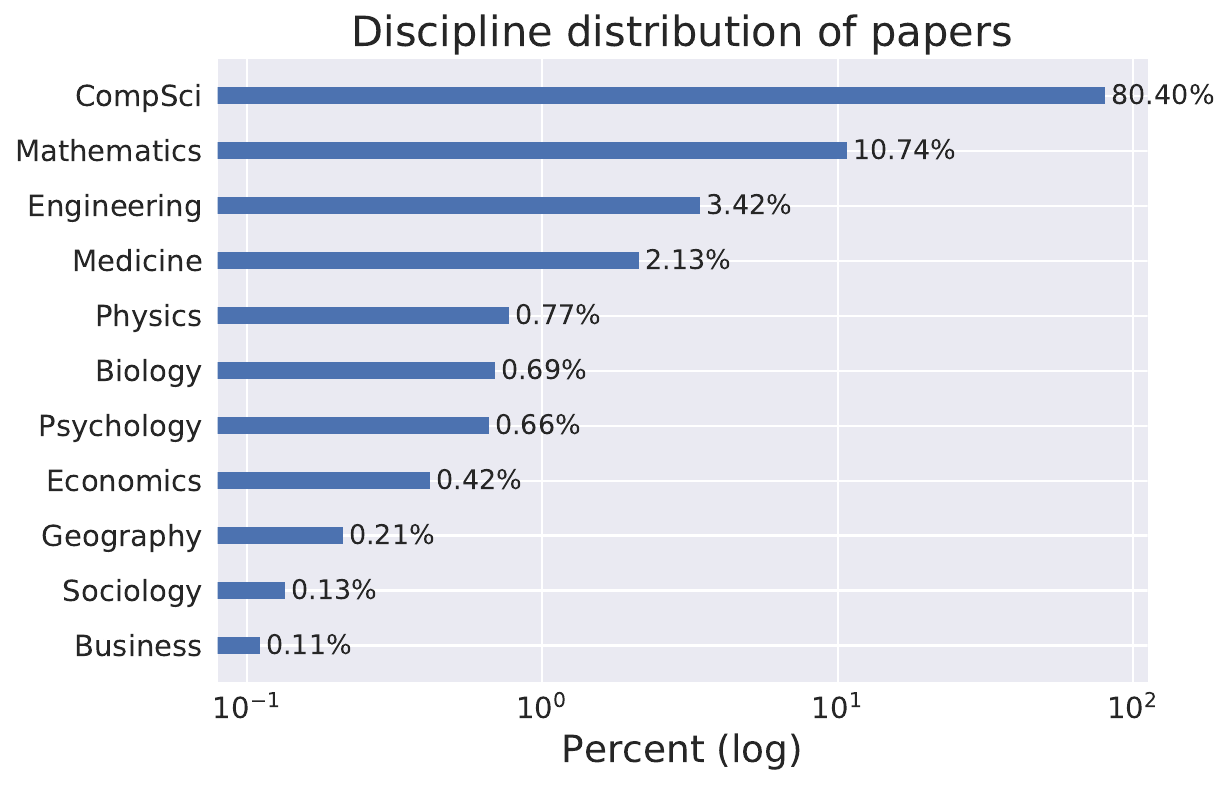}
    \vspace{-0.35cm}
    \upv
    \caption{Discipline distribution of papers in \ourdata. Log scale is used for clearer illustration. Disciplines with lower than 0.1\% distribution are omitted.}
    \label{fig_discipline_dist}
    \downv
\end{figure}

\paragraph{Discipline Analysis}
In Fig.~\ref{fig_discipline_dist}, we show the discipline distribution of papers in \ourdata.
The discipline information is derived from the field of study in Microsoft Academic Graph (MAG) \cite{shen-etal-2018-web}.
We take the top field of study for each paper if there are multiple.
We note that the discipline distribution in \ourdata is quite different from its data source S2ORC \cite{lo-etal-2020-s2orc} where medicine and biology dominate.
In contrast, most papers in \ourdata are in computer science.
The shift in discipline distribution is because we explicitly keep papers with a Related Work section, where around 82.8\% are computer science papers.
We then take the citation texts in the above papers, which largely lead to papers in similar disciplines.
As a result, most papers in \ourdata are from computer science, mathematics, and engineering.

\begin{figure}[t]
    \includegraphics[width=1.\linewidth,center]{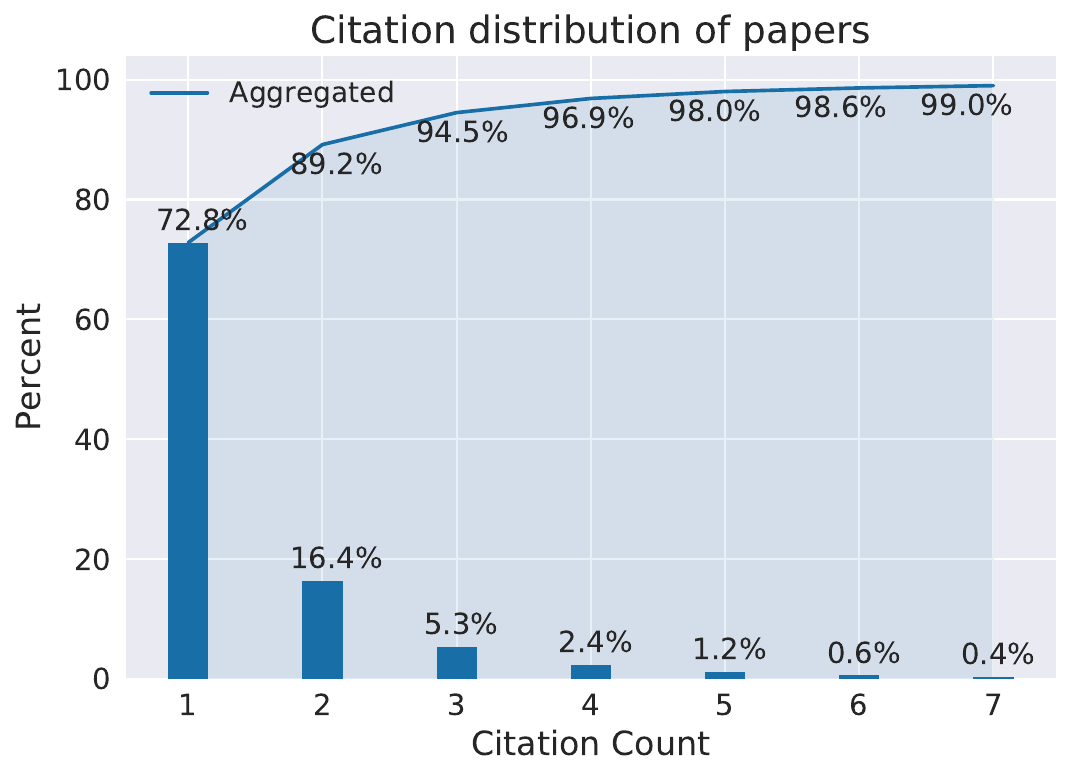}
    \vspace{-0.35cm}
    \upv
    \caption{Citation distribution of papers in \ourdata.}
    \label{fig_citation_dist}
    \downv
\end{figure}

\paragraph{Citation Analysis}
In Fig.~\ref{fig_citation_dist}, we show the average number of citations for papers in \ourdata.
Note that the citation count shown does NOT reflect the total number of citations due to data filtering, but how many times a paper appears in \ourdata as examples (with the same input and different citation sentences as target output).
In total, there are 59,707 unique papers in \ourdata with an average citation of 1.56, and 98\% of the papers have fewer than 5 citations.
Compared to prior work, we do not only target popularly cited papers \cite{yasunaga2019scisummnet} and use different citation texts as different training examples instead of multiple reference summaries~\cite{cachola-etal-2020-tldr}.

\paragraph{Human Evaluation}
We randomly sample 50 examples from \ourdata and ask two human annotators with a background in computer science to examine whether the citation sentences can serve as high-quality summaries of the papers.
Similar to \citet{cachola-etal-2020-tldr}, we use a 4-point scale for evaluation with 1 - false or misleading, 2 - partially accurate, 3 - mostly accurate, and 4 - accurate.
The rating distribution is listed in Table~\ref{table_human_eval}.
80\% citation sentences are considered (mostly) accurate to be used as summaries of the cited papers.
On the other hand, there are still 10\% misleading summaries, which we argue is quite common in automatically created summarization datasets \cite{mao-etal-2020-facet}.
We show 4 examples corresponding to each rating in App.~\ref{sec_human_eval_case}.
We will further verify the quality of \ourdata by adapting models pre-trained on it to new tasks and domains (Sec.~\ref{sec_exp_transfer}).

\begin{table}[ht]
\centering
\resizebox{.77\columnwidth}{!}{

\begin{tabular}{lcccc}
\toprule
\textbf{Rating} & \textbf{1}   & \textbf{2} & \textbf{3} & \textbf{4}  \\ 
\midrule
\textbf{Percentage} & 10\% & 20\% & 28\% & 42\% \\

 \bottomrule
\end{tabular}
}
\upv
\caption{Ratings of citation sentences in \ourdata regarding whether they can serve as high-quality summaries of the cited papers.}
\downv
\label{table_human_eval}
\end{table}

\section{Experiments on \ourdata}
\label{sec_exp_ourdata}
In this section, we experiment on \ourdata with state-of-the-art baselines and analyze their performance under different setups to provide references for future studies.
Implementation and training details are provided in App.~\ref{sec_implement}.

\subsection{Examined Methods}
We use BART-large \cite{lewis-etal-2020-bart} and PEGASUS-large \cite{zhang2020pegasus} as the base models as they are the state-of-the-art methods on multiple summarization datasets.
We examine the base models with different inputs such as paper abstract (Abs), abstract+introduction+conclusion (AIC), and abstract+title.
In addition to using the TLDR (citation text) as the only generation target, we evaluate two multi-task settings with paper title and discipline (Disci)\footnote{Here, we cast discipline classification as a seq2seq task~\cite{mao-etal-2021-reader}. We found that all generated outputs are valid discipline names in our experiments.} as the targets, where different prefix tokens are added to the input such that the model can generate different targets given the same paper abstract as input~\cite{cachola-etal-2020-tldr}.

We further evaluate the following extractive baselines.
\textsc{Ext-Lead}: a method that takes the first sentence of the paper abstract, which performs fairly well in news summarization.
\textsc{Ext-Heuristic}: a heuristic method that looks for the first sentence containing ``propose'', ``introduce'', or ``in this paper'', as such sentences likely reflect the contribution of the paper. It falls back to \textsc{Ext-Lead} if no such sentences are found.
\textsc{Ext-Oracle}: an upper bound that matches each sentence in the paper abstract with the reference summary and takes the sentence with the highest ROUGE-2 F1.

\subsection{Results}
In Table~\ref{table_res_ourdata}, we show the results of various baseline methods on \ourdata.
When given paper abstract as the source document, PEGASUS performs worse than BART and we thus use BART as the major model in the following experiments.
Further adding paper introduction and conclusion to the model input slightly improves model performance, at the expense of longer training time and increased memory usage.
The gains brought by adding title and discipline information to model input are marginal, while using them for multi-task learning does not lead to clearly better results. The fact that methods proposed by recent studies such as multi-task learning \cite{cachola-etal-2020-tldr} perform ineffectively on \ourdata indicates that \ourdata remains an unresolved and challenging scenario. 

For the extractive baselines, \textsc{Ext-Lead} performs significantly worse than that in the news domain \cite{mao-etal-2020-facet}. \textsc{Ext-Heuristic} improves upon \textsc{Ext-Lead} drastically and yet lags behind state-of-the-art methods by a large margin.
\textsc{Ext-Oracle} performs the best, the performance of which is generally consistent with the numbers on the human-annotated SciTLDR dataset \cite{cachola-etal-2020-tldr}.
On the other hand, the fact that abstractive methods have approached the extractive upper bound indicates that more abstraction is needed to further improve model performance on \ourdata. 

We believe that \ourdata provides a well-established testbed for future studies on (scientific) extreme summarization.
The following future directions may be worth exploring: 1) how to better understand the structure and content of scientific papers with domain knowledge (via relevant papers, terminology, taxonomies, etc); 2) how to better capture the differences in writing styles across various domains; and 3) how to improve the saliency, factual correctness, and explainability of TLDR summaries given their conciseness.

\begin{table}[t]
\centering
\resizebox{1.02\columnwidth}{!}{

\begin{tabular}{llccc}
\toprule
\textbf{Method} & & \textbf{R-1} & \textbf{R-2} & \textbf{R-L} \\
\midrule
\textbf{Source} & \textbf{Target} \\
Abs&TLDR    & 41.86        & 19.36        & 33.72       \\
Abs&TLDR (PEGASUS) & 41.56 & 18.63 & 33.45 \\
AIC &TLDR & 41.99 & 19.52 & 33.89\\

Abs+Title &TLDR  & 42.02 & 19.44 & 33.78\\
Abs+Disci &TLDR  & 42.01 & 19.34 & 33.72 \\
Abs&TLDR/Title  & 41.85 & 19.21 & 33.42 \\
Abs&TLDR/Title/Disci & 41.89 & 19.51 & 33.73 \\
\midrule
\multicolumn{2}{l}{\textsc{Ext-Lead}}     & 21.94        & 7.35        & 17.36        \\
\multicolumn{2}{l}{\textsc{Ext-Heuristic}}    & 29.32        & 12.53        & 23.99        \\
\multicolumn{2}{l}{\textsc{Ext-Oracle}}   & {44.17}        & {27.22}       & {38.32}       \\
\bottomrule
\end{tabular}

}
\upv
\caption{Performance of different methods on \ourdata. BART-large is used as the base model if not otherwise specified. ``/'' indicates multi-task learning.  R stands for ROUGE \cite{lin-2004-rouge} in all the tables.}
\downv
\label{table_res_ourdata}
\end{table}

\section{Transferring to New Tasks and Domains with \ours}
\label{sec_exp_transfer}
To further verify the quality and usefulness of \ourdata, we adapt models pre-trained on \ourdata to new tasks and domains, some of which are rather different from \ourdata and make model transfer with limited supervision very challenging.

Specifically, we name our pre-trained model as \ours (\textbf{Ci}tation \textbf{Te}xt-guided \textbf{S}ummarizer).
\ours uses the simplest form in Sec.~\ref{sec_exp_ourdata} with paper abstract as input and TLDR as target output.
We evaluate  \ours on various downstream tasks with no fine-tuning (zero-shot) or limited training examples (few-shot), including scientific extreme summarization on SciTLDR \cite{cachola-etal-2020-tldr}, news extreme summarization on XSum \cite{narayan-etal-2018-dont}, and news headline generation on Gigaword \cite{rush-etal-2015-neural}.
Additionally, we evaluate \ours on two more diverse tasks in the scientific domain,  namely discipline classification and title generation, in a fully-supervised setting.

\subsection{Scientific Extreme Summarization}
\label{sec_transfer_scitldr}

\paragraph{Setup}
SciTLDR \cite{cachola-etal-2020-tldr}, the human-annotated scientific extreme summarization dataset, is an ideal testbed for further verifying the quality and usefulness of \ourdata since they both target extreme summarization, belong to the scientific domain (though \ourdata involves more disciplines), and share similar input/output formats (though \ourdata has slightly longer inputs).
One noticeable difference, however, is the point of view of the summaries -- in SciTLDR the reference summaries typically start with ``We'' or ``This paper'', while in \ourdata they often begin with ``AuthorName et al.'' (replaced by a special token ``REF'' during preprocessing).

We propose two simple techniques to tackle such subtle style differences when adapting \ours to SciTLDR in a zero-shot setting without fine-tuning.
The first technique is \textit{post-processing}: we replace ``REF'' with ``This paper'' if the summary begins with ``REF'' and remove all other ``REF'' within the summary.
The second technique is \textit{prompting}: we use ``This paper'' as a prompt in the model decoder such that the summary always starts with ``This paper''. 
Similarly, in the few-shot setting, we replace the leading ``We'' with ``This paper REF'' in the reference summaries of SciTLDR (on the training set only) to alleviate the style mismatch.

We use BART-large \cite{lewis-etal-2020-bart} as the base model of \ours since most baselines on SciTLDR, including the state-of-the-art methods, use the same base model.

\begin{table}[t]
\centering
\resizebox{1\columnwidth}{!}{

\begin{tabular}{lccc}
\toprule
\textbf{Method}   & \textbf{R-1} & \textbf{R-2} & \textbf{R-L}  \\ \midrule

\textsc{Ext-Oracle}  & \textbf{47.7} & 24.7  & 38.5  \\
\textbf{Fully-supervised}\\
PACSUM \cite{zheng-lapata-2019-sentence}    & 19.3                     & 4.0                  & 15.1                  \\
BERTSum \cite{liu-lapata-2019-text}  & 38.5                     & 16.6                  & 30.5               \\
MatchSum \cite{zhong-etal-2020-extractive}  & 42.7                & 20.0                  & 34.0                   \\
\bart \cite{lewis-etal-2020-bart}      & 43.3                     & 20.8                  & 35.0                 \\
\bartxsum \cite{lewis-etal-2020-bart} & 42.5                     & 21.1                  & 34.9             \\
\Catts \cite{cachola-etal-2020-tldr}               & 43.8         & 20.9                  & 35.5             \\
\Cattsxsum   \cite{cachola-etal-2020-tldr}           & 44.3             & 21.3             & 35.9           \\

\midrule
\textbf{Zero-shot}\\
\cellcolor{na}\ours (post-processing) & 43.4 & 19.7 & 34.8 \\
\cellcolor{na}\ours (prompting) & 43.5 & 20.9& \textbf{36.0}\\
\cellcolor{na}\ours (prompting, gold 3 tokens) & 46.3 & \textbf{27.4}& \textbf{41.9}\\
\midrule
\textbf{Few-shot}\\

\cellcolor{na}\ours (32-shot) & 43.8 & 21.4 & 36.4 \\
BART (128-shot) & 42.1 & 19.1 & 33.4\\
\cellcolor{na}\ours (128-shot) & \textbf{44.5} & \textbf{21.6} & \textbf{36.5} \\

 \bottomrule
\end{tabular}
}
\upv
\caption{Performance comparison on SciTLDR \cite{cachola-etal-2020-tldr} using its official evaluation script.}
\downv
\label{table_res_scitlr}
\end{table}

\paragraph{Zero-shot Results}

In Table \ref{table_res_scitlr}, we show the performance comparison of different methods on SciTLDR.
In the zero-shot setting, \ours (post-processing) outperforms competitive\textit{ fully-supervised} baselines such as BERTSum \cite{liu-lapata-2019-text}.
\ours (prompting) performs even better than \ours (post-processing), outperforming the fully-supervised BART model it is based upon.
Such results demonstrate the benefits of pre-training on \ourdata.
\ours (prompting), without any fine-tuning, is also on par with the state-of-the-art method \Catts \cite{cachola-etal-2020-tldr}, while slightly worse than \Cattsxsum, which pre-trains on the XSum dataset \cite{narayan-etal-2018-dont} first.

We additionally test a zero-shot upper bound for \ours by providing our prompting model with the \textit{first 3 tokens} in the reference summary (the most common ones are ``We propose a'' and ``We present a'') such that it knows how to start to summarize and (hopefully) which aspect to focus on.
\ours (prompting, gold 3 tokens) achieves competitive ROUGE-1 and significantly better ROUGE-2/L than the extractive upper bound \textsc{Ext-Oracle} that has access to the \textit{entire reference summary}.

\paragraph{Few-shot Results}
In the few-shot setting, \ours with 32 examples improves over its zero-shot counterpart.
Furthermore, 128-shot \ours outperforms all fully-supervised methods and achieves new state-of-the-art results on SciTLDR.
In contrast, a 128-shot BART model without first pre-training on \ourdata largely lags behind, performing even worse than our zero-shot \ours.
Such results again show the effectiveness of our pre-training strategy and the quality of \ourdata despite being automatically created thanks to our quality control.

\paragraph{Data Overlap}
To ensure that the superior generalizability of \ours does not merely come from data leakage, we detect the overlap between \ourdata and SciTLDR.
We consider two papers (near) identical if their TF-IDF cosine similarity is greater than 0.9 and find that only 9.7\% papers in the test set of SciTLDR appear in the training set of \ourdata. Also, note that the training labels in \ourdata are automatically extracted citation sentences and different from SciTLDR.

\subsection{Scientific Discipline Classification and Title Generation}
\label{sec_selftransfer}
We have demonstrated the effectiveness of \ours on the task of scientific extreme summarization.
Next, we explore the feasibility of transferring \ours to more diverse tasks.

\paragraph{Setup}
We evaluate \ours with the task of scientific discipline classification and title generation.
Similar to the multi-task experiments in Sec.~\ref{sec_exp_ourdata}, we use the same dataset split and model input, while replacing the generation target from summaries (citation texts) to the discipline or title of the papers.
Examples with unavailable discipline or title are removed.
We use BART-large as the base model for this experiment and compare BART with \ours in an apple-to-apple comparison.

\paragraph{Results}
In Table \ref{table_res_selftransfer}, we show the performance comparison on title generation and discipline classification.
\ours consistently outperforms BART on both tasks, although the differences are not as significant as in other low-resource transfer experiments.
The moderate gains are possibly because there is abundant training data for the two tasks and continuous pre-training thus does not help much.
As evidence, the (unweighted) Macro-F1 of \ours is considerably better than BART, which we found is because \ours performs well on those disciplines with fewer examples.
Regarding the Weighted-F1, \ours is only slightly better as most papers belong to a single discipline (computer science) that dominates the score.

\begin{table}[t]
\centering
\resizebox{1\columnwidth}{!}{

\begin{tabular}{lccccc}
\toprule
   & \multicolumn{3}{c}{\textbf{Title Generation}} & \multicolumn{2}{c}{\textbf{Discipline Classification}} \\ 
    \cmidrule(lr){2-4} \cmidrule(lr){5-6}
\textbf{Method} & \textbf{R-1} & \textbf{R-2} & \textbf{R-L} & \textbf{Macro-F1} & \textbf{Weighted-F1}\\
\midrule
BART     & 52.03        & 30.15      & 45.99  & 0.24 &  0.77   \\
\cellcolor{na}\ours & \textbf{52.50} & \textbf{30.42} & \textbf{46.26} & \textbf{0.30} & \textbf{0.78} \\

\bottomrule
\end{tabular}

}
\upv
\caption{Comparison of \ours and its base model (BART) on title generation and  discipline classification.}
\downv
\label{table_res_selftransfer}
\end{table}

\subsection{News Extreme Summarization}
\label{sec_transfer_xsum}
\paragraph{Setup}
With the success on different tasks in the scientific domain, we next evaluate \ours on a more difficult setting where the domain is significantly different while the task is still extreme summarization.
We take the XSum dataset \cite{narayan-etal-2018-dont} in the news domain for this purpose.
We mainly use PEGASUS-large \cite{zhang2020pegasus} as the base model of \ours as its fully-supervised version holds the state-of-the-art results on XSum.
We additionally evaluate \oursTitle in the zero-shot setting, which is the variant used for title generation in Sec.~\ref{sec_selftransfer}.

\paragraph{Zero-shot Results}
In Table~\ref{table_res_xsum}, we show the results on XSum with various training data sizes.
In the zero-shot setting, \ours significantly improves over its base model PEGASUS (+7.2 ROUGE-1).
In addition, \ours is on par with other pre-trained models such as BART-LB and T5-LB \cite{zhu2021leveraging}, which are specifically designed to leverage the lead bias in the news domain and require much more resources (32 vs. 1 GPU, 21.4M vs. 83K training examples) for summarization pre-training.
\oursTitle further improves over \ours and outperforms most zero-shot baselines (+8.9 ROUGE-1 over PEGASUS).
\oursTitle does not outperform WikiTransfer \cite{fabbri-etal-2021-improving}, which is somewhat expected as WikiTransfer carefully prepares its pre-training data to specific downstream tasks given, \eg, summary length and its level of abstraction.
Unlike WikiTransfer \cite{fabbri-etal-2021-improving}, \ours does not involve any downstream task-specific data selection or model tuning -- we use the same \ourdata corpus in all the experiments.

\begin{table}[t]
\centering
\resizebox{1\columnwidth}{!}{

\begin{tabular}{lccc}
\toprule

\textbf{Method}   & \textbf{R-1} & \textbf{R-2} & \textbf{R-L}  \\ 
\midrule
\textbf{Fully-supervised}\\
\textsc{PtGen} \cite{see-etal-2017-get} &29.70	&9.21	&23.24 \\
BERTSum \cite{liu-lapata-2019-text} &38.81	&16.50	&31.27 \\
BART \cite{lewis-etal-2020-bart} & 45.14	&22.27	&37.25 \\
PEGASUS \cite{zhang2020pegasus} &\textbf{47.21}	&\textbf{24.56}	&\textbf{39.25} \\

\midrule
\textbf{Zero-shot}\\
BART \cite{lewis-etal-2020-bart}  &  15.40 & 2.63 & 10.74 \\
PEGASUS \cite{zhang2020pegasus} & 19.27&3.00&12.72  \\
T5-LB  \cite{zhu2021leveraging} & 26.06 & 6.77 & 20.47\\
BART-LB  \cite{zhu2021leveraging} &{26.18} & {7.60} & {20.92} \\

WikiTransfer \cite{fabbri-etal-2021-improving} &\textbf{31.85 }& \textbf{10.44} & \textbf{23.75} \\ 
\cellcolor{na}\ours & {26.43} & {7.17} & {19.64}   \\
\cellcolor{na}\oursTitle & {28.21} & {8.40} & {21.81}   \\

\midrule
\textbf{10-shot}\\
BART \cite{lewis-etal-2020-bart}  & {31.34} & {9.98} & {23.44} \\
PEGASUS\cite{zhang2020pegasus} & 19.39 & 3.45 & 14.02\\
WikiTransfer \cite{fabbri-etal-2021-improving} & 35.17 &12.76 &26.80 \\ 
\cellcolor{na}\ours &  \textbf{36.21} & \textbf{14.22} & \textbf{28.18} \\

\midrule
\textbf{100-shot}\\
BART \cite{lewis-etal-2020-bart} & 34.16&12.62&26.66 \\
PEGASUS \cite{zhang2020pegasus}& 39.07&16.44&31.27 \\
WikiTransfer \cite{fabbri-etal-2021-improving}  & 37.26 &14.20 &28.85 \\
\cellcolor{na}\ours & \textbf{41.45} & \textbf{18.74} & \textbf{33.29} \\

 \bottomrule

\end{tabular}
}
\upv
\caption{Performance comparison on the XSum dataset. Our few-shot results are averaged over 3 runs.}
\downv
\label{table_res_xsum}
\end{table}

\paragraph{Few-shot Results}
When given a few examples for fine-tuning, \ours quickly adapts to the new task despite the domain mismatch during pre-training.
We observe that \ours consistently outperforms not only its base model  but all other baseline methods, including WikiTransfer, and achieves state-of-the-art few-shot performance on XSum.
In particular, \ours performs better than fully-supervised methods such as BERTSum \cite{liu-lapata-2019-text} with only 100 examples.

\subsection{News Headline Generation}
\paragraph{Setup}
To take a step further, we study the transfer performance of \ours to news headline generation.
We use the Gigaword headline generation dataset \cite{rush-etal-2015-neural} for this evaluation.
We again consider two variants of \ours, one pre-trained with citation texts as the generation target and the other further pretrained with paper titles as in Sec.~\ref{sec_selftransfer}.
We use BART-large \cite{lewis-etal-2020-bart} as the base model in this evaluation.

\paragraph{Results}
In Table~\ref{table_res_gigaword}, we show the results of various methods on news headline generation.
\ours again outperforms its base model (BART) significantly and achieves competitive performance with most unsupervised and zero-shot methods designed for news summarization \cite{zhang2020pegasus,zhu2021leveraging}.
\oursTitle further achieves state-of-the-art zero-shot performance despite pre-training on the scientific domain, demonstrating the generalizability and usefulness of \ourdata.

\begin{table}[t]
\centering
\resizebox{1\columnwidth}{!}{

\begin{tabular}{lccc}
\toprule
\textbf{Method}   & \textbf{R-1} & \textbf{R-2} & \textbf{R-L}  \\ \midrule

\textbf{Fully-supervised}\\
PEGASUS \cite{zhang2020pegasus} &{39.12}	&{19.86}	&{36.24} \\
\midrule

\textbf{Unsupervised}\\
SEQ$^3$ \cite{baziotis-etal-2019-seq} & 25.39 & 8.21 & 22.68\\
Brief \cite{wang-lee-2018-learning} & 21.26 & 5.60 & 18.89\\
TED \cite{yang-etal-2020-ted} &25.58 & 8.94 & 22.83\\
\midrule

\textbf{Zero-shot}\\
T5 \cite{raffel2020exploring} & 15.67 & 4.86 & 14.38\\
BART \cite{lewis-etal-2020-bart} & 22.07 & 7.47 & 20.02\\
PEGASUS \cite{zhang2020pegasus} & 23.39 & 7.59 & 20.20\\
T5-LB \cite{zhu2021leveraging} & 24.00 & 8.19 & 21.62\\
BART-LB \cite{zhu2021leveraging} & 25.14 & 8.72 & 22.35 \\

\cellcolor{na}\ours & {24.75} & {8.42} & {21.84} \\
\cellcolor{na}\oursTitle & \textbf{27.87} & \textbf{10.43} & \textbf{24.56} \\

 \bottomrule
\end{tabular}
}
\upv
\caption{Performance comparison on the Gigaword news headline generation dataset \cite{rush-etal-2015-neural}.}
\downv
\label{table_res_gigaword}
\end{table}

\section{Related Work}

\paragraph{Citation Text Generation}
There have been prior studies utilizing citation texts for different purposes.
One popular line of work focuses on the generation of the citation texts for writing assistance or paper comparison \cite{xing-etal-2020-automatic,luu-etal-2021-explaining,chen-etal-2021-capturing,ge-etal-2021-baco}.
However, they typically do not distinguish the citation texts that can serve as summaries of the cited paper from those used for other purposes, \eg, background or result comparison \cite{cohan-etal-2019-structural}.
For example, \citet{chen-etal-2021-capturing} treat citation text generation as a multi-document summarization task, where the target output is a paragraph with more than two citations and the model input is the abstracts of all cited papers.
There is no filtering regarding the citation texts and all the paragraphs with enough citations are included.
Besides including citation texts with various intents and the lack of quality control, prior studies differ from \ourdata in that they target longer outputs, \eg, multiple sentences \cite{xing-etal-2020-automatic} or the entire Related Work section \cite{lu-etal-2020-multi-xscience,chen-etal-2021-capturing}.

\paragraph{Citation Text for Paper Summarization}
Another line of work does not generate but extracts the citation texts and either uses them to form a summary directly \cite{nakov2004citances,abu-jbara-radev-2011-coherent,qazvinian2013generating} or treats them as a bridge to the cited paper \cite{cohan-goharian-2015-scientific,yasunaga2019scisummnet}.
Specifically, the citation texts in the latter studies are used to find relevant contexts in the cited paper (called \textit{citation contexts}).
Then, a \textit{long} summary is formulated primarily using the cited paper, \eg, by selecting sentences from the citation contexts \cite{cohan-goharian-2015-scientific}.
Unlike \ours, prior citation-based summarization methods require (often multiple) citation texts of a paper as \textit{input}, which are unavailable for new papers. In addition, they do not target ultra-short but abstract-long summaries.

\paragraph{Extreme Summarization}
Extreme summarization aims to form ultra-short summaries of the documents.
Notable benchmarks in this direction include XSum \cite{narayan-etal-2018-dont} and NewSHead \cite{headline2020} in the news domain, SciTLDR \cite{cachola-etal-2020-tldr} in the scientific domain, and Webis-TLDR-17 \cite{volske-etal-2017-tl} for social media summarization.
Compared to SciTLDR, our \ourdata dataset is significantly larger in scale, from more venues than OpenReview, and composed of various disciplines.

\paragraph{Summarization with Limited Supervision}
Our work is also related to unsupervised and zero/few-shot summarization that constructs weakly supervised guidance signals using e.g., data characteristics \cite{chu2019meansum,mao-etal-2020-multi}, domain knowledge \cite{zhu2021sumdocs}, or pseudo labeled data \cite{yang-etal-2020-ted,zhong2022unsupervised}. 
Compared to prior studies, \ours shows great cross-domain capability that has not been well explored.

\section{Conclusion}
In this paper, we propose a simple yet effective approach to automatically extracting ultra-short paper summaries from citation texts.
Based on the proposed approach, we create a large-scale, high-quality benchmark for scientific extreme summarization.
We conduct a comprehensive analysis on the created benchmark and further demonstrate that models pre-trained on it exhibit superior generalizability to new tasks and domains such as news extreme summarization and headline generation with limited supervision.

\section*{Limitations}
Regarding data collection, while we have taken multiple steps to improve data quality, as in all automatically created datasets, there are still low-quality examples. We show some examples of low quality in App.~\ref{sec_human_eval_case}.
Limiting citation texts to the Related Work section improves data quality, but also excludes the majority of available citation sentences and makes \ourdata concentrated in the field of computer science and engineering.

Regarding model performance, our transfer experiments are performed in scientific and news domains. While promising, there is no guarantee that \ours works well in other domains.
Also, with abundant in-domain training data, pre-training on \ourdata may not lead to significant improvements.

\bibliography{anthology,custom}

\begin{thebibliography}{42}
\expandafter\ifx\csname natexlab\endcsname\relax\def\natexlab#1{#1}\fi

\bibitem[{Abu-Jbara and Radev(2011)}]{abu-jbara-radev-2011-coherent}
Amjad Abu-Jbara and Dragomir Radev. 2011.
\newblock \href {https://aclanthology.org/P11-1051} {Coherent citation-based
  summarization of scientific papers}.
\newblock In \emph{Proceedings of the 49th Annual Meeting of the Association
  for Computational Linguistics: Human Language Technologies}, pages 500--509,
  Portland, Oregon, USA. Association for Computational Linguistics.

\bibitem[{Baziotis et~al.(2019)Baziotis, Androutsopoulos, Konstas, and
  Potamianos}]{baziotis-etal-2019-seq}
Christos Baziotis, Ion Androutsopoulos, Ioannis Konstas, and Alexandros
  Potamianos. 2019.
\newblock \href {https://doi.org/10.18653/v1/N19-1071} {{SEQ}{\^{}}3:
  Differentiable sequence-to-sequence-to-sequence autoencoder for unsupervised
  abstractive sentence compression}.
\newblock In \emph{Proceedings of the 2019 Conference of the North {A}merican
  Chapter of the Association for Computational Linguistics: Human Language
  Technologies, Volume 1 (Long and Short Papers)}, pages 673--681, Minneapolis,
  Minnesota. Association for Computational Linguistics.

\bibitem[{Cachola et~al.(2020)Cachola, Lo, Cohan, and
  Weld}]{cachola-etal-2020-tldr}
Isabel Cachola, Kyle Lo, Arman Cohan, and Daniel Weld. 2020.
\newblock \href {https://doi.org/10.18653/v1/2020.findings-emnlp.428} {{TLDR}:
  Extreme summarization of scientific documents}.
\newblock In \emph{Findings of the Association for Computational Linguistics:
  EMNLP 2020}, pages 4766--4777, Online. Association for Computational
  Linguistics.

\bibitem[{Chen et~al.(2021)Chen, Alamro, Li, Gao, Zhang, Zhao, and
  Yan}]{chen-etal-2021-capturing}
Xiuying Chen, Hind Alamro, Mingzhe Li, Shen Gao, Xiangliang Zhang, Dongyan
  Zhao, and Rui Yan. 2021.
\newblock \href {https://doi.org/10.18653/v1/2021.acl-long.473} {Capturing
  relations between scientific papers: An abstractive model for related work
  section generation}.
\newblock In \emph{Proceedings of the 59th Annual Meeting of the Association
  for Computational Linguistics and the 11th International Joint Conference on
  Natural Language Processing (Volume 1: Long Papers)}, pages 6068--6077,
  Online. Association for Computational Linguistics.

\bibitem[{Chu and Liu(2019)}]{chu2019meansum}
Eric Chu and Peter Liu. 2019.
\newblock Meansum: a neural model for unsupervised multi-document abstractive
  summarization.
\newblock In \emph{International Conference on Machine Learning}, pages
  1223--1232. PMLR.

\bibitem[{Cohan et~al.(2019)Cohan, Ammar, van Zuylen, and
  Cady}]{cohan-etal-2019-structural}
Arman Cohan, Waleed Ammar, Madeleine van Zuylen, and Field Cady. 2019.
\newblock \href {https://doi.org/10.18653/v1/N19-1361} {Structural scaffolds
  for citation intent classification in scientific publications}.
\newblock In \emph{Proceedings of the 2019 Conference of the North {A}merican
  Chapter of the Association for Computational Linguistics: Human Language
  Technologies, Volume 1 (Long and Short Papers)}, pages 3586--3596,
  Minneapolis, Minnesota. Association for Computational Linguistics.

\bibitem[{Cohan et~al.(2018)Cohan, Dernoncourt, Kim, Bui, Kim, Chang, and
  Goharian}]{cohan-etal-2018-discourse}
Arman Cohan, Franck Dernoncourt, Doo~Soon Kim, Trung Bui, Seokhwan Kim, Walter
  Chang, and Nazli Goharian. 2018.
\newblock \href {https://doi.org/10.18653/v1/N18-2097} {A discourse-aware
  attention model for abstractive summarization of long documents}.
\newblock In \emph{Proceedings of the 2018 Conference of the North {A}merican
  Chapter of the Association for Computational Linguistics: Human Language
  Technologies, Volume 2 (Short Papers)}, pages 615--621, New Orleans,
  Louisiana. Association for Computational Linguistics.

\bibitem[{Cohan and Goharian(2015)}]{cohan-goharian-2015-scientific}
Arman Cohan and Nazli Goharian. 2015.
\newblock \href {https://doi.org/10.18653/v1/D15-1045} {Scientific article
  summarization using citation-context and article{'}s discourse structure}.
\newblock In \emph{Proceedings of the 2015 Conference on Empirical Methods in
  Natural Language Processing}, pages 390--400, Lisbon, Portugal. Association
  for Computational Linguistics.

\bibitem[{Fabbri et~al.(2021)Fabbri, Han, Li, Li, Ghazvininejad, Joty, Radev,
  and Mehdad}]{fabbri-etal-2021-improving}
Alexander Fabbri, Simeng Han, Haoyuan Li, Haoran Li, Marjan Ghazvininejad,
  Shafiq Joty, Dragomir Radev, and Yashar Mehdad. 2021.
\newblock \href {https://doi.org/10.18653/v1/2021.naacl-main.57} {Improving
  zero and few-shot abstractive summarization with intermediate fine-tuning and
  data augmentation}.
\newblock In \emph{Proceedings of the 2021 Conference of the North American
  Chapter of the Association for Computational Linguistics: Human Language
  Technologies}, pages 704--717, Online. Association for Computational
  Linguistics.

\bibitem[{Ge et~al.(2021)Ge, Dinh, Liu, Su, Lu, Wang, and
  Diesner}]{ge-etal-2021-baco}
Yubin Ge, Ly~Dinh, Xiaofeng Liu, Jinsong Su, Ziyao Lu, Ante Wang, and Jana
  Diesner. 2021.
\newblock \href {https://doi.org/10.18653/v1/2021.acl-long.116} {{BACO}: A
  background knowledge- and content-based framework for citing sentence
  generation}.
\newblock In \emph{Proceedings of the 59th Annual Meeting of the Association
  for Computational Linguistics and the 11th International Joint Conference on
  Natural Language Processing (Volume 1: Long Papers)}, pages 1466--1478,
  Online. Association for Computational Linguistics.

\bibitem[{Gu et~al.(2020)Gu, Mao, Han, Liu, Yu, Wu, Yu, Finnie, Zhai, and
  Zukoski}]{headline2020}
Xiaotao Gu, Yuning Mao, Jiawei Han, Jialu Liu, Hongkun Yu, You Wu, Cong Yu,
  Daniel Finnie, Jiaqi Zhai, and Nicholas Zukoski. 2020.
\newblock {Generating Representative Headlines for News Stories}.
\newblock In \emph{Proc. of the the Web Conf. 2020}.

\bibitem[{Jaidka et~al.(2016)Jaidka, Kumar~Chandrasekaran, Rustagi, and
  Kan}]{jaidka-etal-2016-overview}
Kokil Jaidka, Muthu Kumar~Chandrasekaran, Sajal Rustagi, and Min-Yen Kan. 2016.
\newblock \href {https://aclanthology.org/W16-1511} {Overview of the
  {CL}-{S}ci{S}umm 2016 shared task}.
\newblock In \emph{Proceedings of the Joint Workshop on Bibliometric-enhanced
  Information Retrieval and Natural Language Processing for Digital Libraries
  ({BIRNDL})}, pages 93--102.

\bibitem[{Lev et~al.(2019)Lev, Shmueli-Scheuer, Herzig, Jerbi, and
  Konopnicki}]{lev-etal-2019-talksumm}
Guy Lev, Michal Shmueli-Scheuer, Jonathan Herzig, Achiya Jerbi, and David
  Konopnicki. 2019.
\newblock \href {https://doi.org/10.18653/v1/P19-1204} {{T}alk{S}umm: A dataset
  and scalable annotation method for scientific paper summarization based on
  conference talks}.
\newblock In \emph{Proceedings of the 57th Annual Meeting of the Association
  for Computational Linguistics}, pages 2125--2131, Florence, Italy.
  Association for Computational Linguistics.

\bibitem[{Lewis et~al.(2020)Lewis, Liu, Goyal, Ghazvininejad, Mohamed, Levy,
  Stoyanov, and Zettlemoyer}]{lewis-etal-2020-bart}
Mike Lewis, Yinhan Liu, Naman Goyal, Marjan Ghazvininejad, Abdelrahman Mohamed,
  Omer Levy, Veselin Stoyanov, and Luke Zettlemoyer. 2020.
\newblock \href {https://doi.org/10.18653/v1/2020.acl-main.703} {{BART}:
  Denoising sequence-to-sequence pre-training for natural language generation,
  translation, and comprehension}.
\newblock In \emph{Proceedings of the 58th Annual Meeting of the Association
  for Computational Linguistics}, pages 7871--7880, Online. Association for
  Computational Linguistics.

\bibitem[{Lin(2004)}]{lin-2004-rouge}
Chin-Yew Lin. 2004.
\newblock \href {https://aclanthology.org/W04-1013} {{ROUGE}: A package for
  automatic evaluation of summaries}.
\newblock In \emph{Text Summarization Branches Out}, pages 74--81, Barcelona,
  Spain. Association for Computational Linguistics.

\bibitem[{Liu and Lapata(2019)}]{liu-lapata-2019-text}
Yang Liu and Mirella Lapata. 2019.
\newblock \href {https://doi.org/10.18653/v1/D19-1387} {Text summarization with
  pretrained encoders}.
\newblock In \emph{Proceedings of the 2019 Conference on Empirical Methods in
  Natural Language Processing and the 9th International Joint Conference on
  Natural Language Processing (EMNLP-IJCNLP)}, pages 3730--3740, Hong Kong,
  China. Association for Computational Linguistics.

\bibitem[{Lo et~al.(2020)Lo, Wang, Neumann, Kinney, and
  Weld}]{lo-etal-2020-s2orc}
Kyle Lo, Lucy~Lu Wang, Mark Neumann, Rodney Kinney, and Daniel Weld. 2020.
\newblock \href {https://doi.org/10.18653/v1/2020.acl-main.447} {{S}2{ORC}: The
  semantic scholar open research corpus}.
\newblock In \emph{Proceedings of the 58th Annual Meeting of the Association
  for Computational Linguistics}, pages 4969--4983, Online. Association for
  Computational Linguistics.

\bibitem[{Lu et~al.(2020)Lu, Dong, and Charlin}]{lu-etal-2020-multi-xscience}
Yao Lu, Yue Dong, and Laurent Charlin. 2020.
\newblock \href {https://doi.org/10.18653/v1/2020.emnlp-main.648}
  {Multi-{XS}cience: A large-scale dataset for extreme multi-document
  summarization of scientific articles}.
\newblock In \emph{Proceedings of the 2020 Conference on Empirical Methods in
  Natural Language Processing (EMNLP)}, pages 8068--8074, Online. Association
  for Computational Linguistics.

\bibitem[{Luu et~al.(2021)Luu, Wu, Koncel-Kedziorski, Lo, Cachola, and
  Smith}]{luu-etal-2021-explaining}
Kelvin Luu, Xinyi Wu, Rik Koncel-Kedziorski, Kyle Lo, Isabel Cachola, and
  Noah~A. Smith. 2021.
\newblock \href {https://doi.org/10.18653/v1/2021.acl-long.166} {Explaining
  relationships between scientific documents}.
\newblock In \emph{Proceedings of the 59th Annual Meeting of the Association
  for Computational Linguistics and the 11th International Joint Conference on
  Natural Language Processing (Volume 1: Long Papers)}, pages 2130--2144,
  Online. Association for Computational Linguistics.

\bibitem[{Mao et~al.(2021)Mao, He, Liu, Shen, Gao, Han, and
  Chen}]{mao-etal-2021-reader}
Yuning Mao, Pengcheng He, Xiaodong Liu, Yelong Shen, Jianfeng Gao, Jiawei Han,
  and Weizhu Chen. 2021.
\newblock \href {https://doi.org/10.18653/v1/2021.findings-acl.29}
  {Reader-guided passage reranking for open-domain question answering}.
\newblock In \emph{Findings of the Association for Computational Linguistics:
  ACL-IJCNLP 2021}, pages 344--350, Online. Association for Computational
  Linguistics.

\bibitem[{Mao et~al.(2020{\natexlab{a}})Mao, Liu, Zhu, Ren, and
  Han}]{mao-etal-2020-facet}
Yuning Mao, Liyuan Liu, Qi~Zhu, Xiang Ren, and Jiawei Han. 2020{\natexlab{a}}.
\newblock \href {https://doi.org/10.18653/v1/2020.acl-main.445} {Facet-aware
  evaluation for extractive summarization}.
\newblock In \emph{Proceedings of the 58th Annual Meeting of the Association
  for Computational Linguistics}, pages 4941--4957, Online. Association for
  Computational Linguistics.

\bibitem[{Mao et~al.(2020{\natexlab{b}})Mao, Qu, Xie, Ren, and
  Han}]{mao-etal-2020-multi}
Yuning Mao, Yanru Qu, Yiqing Xie, Xiang Ren, and Jiawei Han.
  2020{\natexlab{b}}.
\newblock \href {https://doi.org/10.18653/v1/2020.emnlp-main.136}
  {Multi-document summarization with maximal marginal relevance-guided
  reinforcement learning}.
\newblock In \emph{Proceedings of the 2020 Conference on Empirical Methods in
  Natural Language Processing (EMNLP)}, pages 1737--1751, Online. Association
  for Computational Linguistics.

\bibitem[{Nakov et~al.(2004)Nakov, Schwartz, Hearst et~al.}]{nakov2004citances}
Preslav~I Nakov, Ariel~S Schwartz, Marti Hearst, et~al. 2004.
\newblock Citances: Citation sentences for semantic analysis of bioscience
  text.
\newblock In \emph{Proceedings of the SIGIR}, volume~4, pages 81--88. Citeseer.

\bibitem[{Narayan et~al.(2018)Narayan, Cohen, and
  Lapata}]{narayan-etal-2018-dont}
Shashi Narayan, Shay~B. Cohen, and Mirella Lapata. 2018.
\newblock \href {https://doi.org/10.18653/v1/D18-1206} {Don{'}t give me the
  details, just the summary! topic-aware convolutional neural networks for
  extreme summarization}.
\newblock In \emph{Proceedings of the 2018 Conference on Empirical Methods in
  Natural Language Processing}, pages 1797--1807, Brussels, Belgium.
  Association for Computational Linguistics.

\bibitem[{Qazvinian et~al.(2013)Qazvinian, Radev, Mohammad, Dorr, Zajic,
  Whidby, and Moon}]{qazvinian2013generating}
Vahed Qazvinian, Dragomir~R Radev, Saif~M Mohammad, Bonnie Dorr, David Zajic,
  Michael Whidby, and Taesun Moon. 2013.
\newblock Generating extractive summaries of scientific paradigms.
\newblock \emph{Journal of Artificial Intelligence Research}, 46:165--201.

\bibitem[{Raffel et~al.(2020)Raffel, Shazeer, Roberts, Lee, Narang, Matena,
  Zhou, Li, and Liu}]{raffel2020exploring}
Colin Raffel, Noam Shazeer, Adam Roberts, Katherine Lee, Sharan Narang, Michael
  Matena, Yanqi Zhou, Wei Li, and Peter~J Liu. 2020.
\newblock Exploring the limits of transfer learning with a unified text-to-text
  transformer.
\newblock \emph{Journal of Machine Learning Research}, 21:1--67.

\bibitem[{Rush et~al.(2015)Rush, Chopra, and Weston}]{rush-etal-2015-neural}
Alexander~M. Rush, Sumit Chopra, and Jason Weston. 2015.
\newblock \href {https://doi.org/10.18653/v1/D15-1044} {A neural attention
  model for abstractive sentence summarization}.
\newblock In \emph{Proceedings of the 2015 Conference on Empirical Methods in
  Natural Language Processing}, pages 379--389, Lisbon, Portugal. Association
  for Computational Linguistics.

\bibitem[{See et~al.(2017)See, Liu, and Manning}]{see-etal-2017-get}
Abigail See, Peter~J. Liu, and Christopher~D. Manning. 2017.
\newblock \href {https://doi.org/10.18653/v1/P17-1099} {Get to the point:
  Summarization with pointer-generator networks}.
\newblock In \emph{Proceedings of the 55th Annual Meeting of the Association
  for Computational Linguistics (Volume 1: Long Papers)}, pages 1073--1083,
  Vancouver, Canada. Association for Computational Linguistics.

\bibitem[{Shen et~al.(2018)Shen, Ma, and Wang}]{shen-etal-2018-web}
Zhihong Shen, Hao Ma, and Kuansan Wang. 2018.
\newblock \href {https://doi.org/10.18653/v1/P18-4015} {A web-scale system for
  scientific knowledge exploration}.
\newblock In \emph{Proceedings of {ACL} 2018, System Demonstrations}, pages
  87--92, Melbourne, Australia. Association for Computational Linguistics.

\bibitem[{V{\"o}lske et~al.(2017)V{\"o}lske, Potthast, Syed, and
  Stein}]{volske-etal-2017-tl}
Michael V{\"o}lske, Martin Potthast, Shahbaz Syed, and Benno Stein. 2017.
\newblock \href {https://doi.org/10.18653/v1/W17-4508} {{TL};{DR}: Mining
  {R}eddit to learn automatic summarization}.
\newblock In \emph{Proceedings of the Workshop on New Frontiers in
  Summarization}, pages 59--63, Copenhagen, Denmark. Association for
  Computational Linguistics.

\bibitem[{Wang and Lee(2018)}]{wang-lee-2018-learning}
Yaushian Wang and Hung-Yi Lee. 2018.
\newblock \href {https://doi.org/10.18653/v1/D18-1451} {Learning to encode text
  as human-readable summaries using generative adversarial networks}.
\newblock In \emph{Proceedings of the 2018 Conference on Empirical Methods in
  Natural Language Processing}, pages 4187--4195, Brussels, Belgium.
  Association for Computational Linguistics.

\bibitem[{Wolf et~al.(2020)Wolf, Debut, Sanh, Chaumond, Delangue, Moi, Cistac,
  Rault, Louf, Funtowicz, Davison, Shleifer, von Platen, Ma, Jernite, Plu, Xu,
  Le~Scao, Gugger, Drame, Lhoest, and Rush}]{wolf-etal-2020-transformers}
Thomas Wolf, Lysandre Debut, Victor Sanh, Julien Chaumond, Clement Delangue,
  Anthony Moi, Pierric Cistac, Tim Rault, Remi Louf, Morgan Funtowicz, Joe
  Davison, Sam Shleifer, Patrick von Platen, Clara Ma, Yacine Jernite, Julien
  Plu, Canwen Xu, Teven Le~Scao, Sylvain Gugger, Mariama Drame, Quentin Lhoest,
  and Alexander Rush. 2020.
\newblock \href {https://doi.org/10.18653/v1/2020.emnlp-demos.6} {Transformers:
  State-of-the-art natural language processing}.
\newblock In \emph{Proceedings of the 2020 Conference on Empirical Methods in
  Natural Language Processing: System Demonstrations}, pages 38--45, Online.
  Association for Computational Linguistics.

\bibitem[{Xing et~al.(2020)Xing, Fan, and Wan}]{xing-etal-2020-automatic}
Xinyu Xing, Xiaosheng Fan, and Xiaojun Wan. 2020.
\newblock \href {https://doi.org/10.18653/v1/2020.acl-main.550} {Automatic
  generation of citation texts in scholarly papers: A pilot study}.
\newblock In \emph{Proceedings of the 58th Annual Meeting of the Association
  for Computational Linguistics}, pages 6181--6190, Online. Association for
  Computational Linguistics.

\bibitem[{Yang et~al.(2020)Yang, Zhu, Gmyr, Zeng, Huang, and
  Darve}]{yang-etal-2020-ted}
Ziyi Yang, Chenguang Zhu, Robert Gmyr, Michael Zeng, Xuedong Huang, and Eric
  Darve. 2020.
\newblock \href {https://doi.org/10.18653/v1/2020.findings-emnlp.168} {{TED}: A
  pretrained unsupervised summarization model with theme modeling and
  denoising}.
\newblock In \emph{Findings of the Association for Computational Linguistics:
  EMNLP 2020}, pages 1865--1874, Online. Association for Computational
  Linguistics.

\bibitem[{Yasunaga et~al.(2019)Yasunaga, Kasai, Zhang, Fabbri, Li, Friedman,
  and Radev}]{yasunaga2019scisummnet}
Michihiro Yasunaga, Jungo Kasai, Rui Zhang, Alexander~R Fabbri, Irene Li, Dan
  Friedman, and Dragomir~R Radev. 2019.
\newblock Scisummnet: A large annotated corpus and content-impact models for
  scientific paper summarization with citation networks.
\newblock In \emph{Proceedings of the AAAI Conference on Artificial
  Intelligence}, volume~33, pages 7386--7393.

\bibitem[{Zhang et~al.(2020)Zhang, Zhao, Saleh, and Liu}]{zhang2020pegasus}
Jingqing Zhang, Yao Zhao, Mohammad Saleh, and Peter Liu. 2020.
\newblock Pegasus: Pre-training with extracted gap-sentences for abstractive
  summarization.
\newblock In \emph{International Conference on Machine Learning}, pages
  11328--11339. PMLR.

\bibitem[{Zhang et~al.(2019)Zhang, Kishore, Wu, Weinberger, and
  Artzi}]{zhang2019bertscore}
Tianyi Zhang, Varsha Kishore, Felix Wu, Kilian~Q Weinberger, and Yoav Artzi.
  2019.
\newblock Bertscore: Evaluating text generation with bert.
\newblock In \emph{International Conference on Learning Representations}.

\bibitem[{Zheng and Lapata(2019)}]{zheng-lapata-2019-sentence}
Hao Zheng and Mirella Lapata. 2019.
\newblock \href {https://doi.org/10.18653/v1/P19-1628} {Sentence centrality
  revisited for unsupervised summarization}.
\newblock In \emph{Proceedings of the 57th Annual Meeting of the Association
  for Computational Linguistics}, pages 6236--6247, Florence, Italy.
  Association for Computational Linguistics.

\bibitem[{Zhong et~al.(2020)Zhong, Liu, Chen, Wang, Qiu, and
  Huang}]{zhong-etal-2020-extractive}
Ming Zhong, Pengfei Liu, Yiran Chen, Danqing Wang, Xipeng Qiu, and Xuanjing
  Huang. 2020.
\newblock \href {https://doi.org/10.18653/v1/2020.acl-main.552} {Extractive
  summarization as text matching}.
\newblock In \emph{Proceedings of the 58th Annual Meeting of the Association
  for Computational Linguistics}, pages 6197--6208, Online. Association for
  Computational Linguistics.

\bibitem[{Zhong et~al.(2022)Zhong, Liu, Ge, Mao, Jiao, Zhang, Xu, Zhu, Zeng,
  and Han}]{zhong2022unsupervised}
Ming Zhong, Yang Liu, Suyu Ge, Yuning Mao, Yizhu Jiao, Xingxing Zhang, Yichong
  Xu, Chenguang Zhu, Michael Zeng, and Jiawei Han. 2022.
\newblock Unsupervised summarization with customized granularities.
\newblock In \emph{Findings of the Association for Computational Linguistics:
  EMNLP 2022}.

\bibitem[{Zhu et~al.(2021{\natexlab{a}})Zhu, Yang, Gmyr, Zeng, and
  Huang}]{zhu2021leveraging}
Chenguang Zhu, Ziyi Yang, Robert Gmyr, Michael Zeng, and Xuedong Huang.
  2021{\natexlab{a}}.
\newblock Leveraging lead bias for zero-shot abstractive news summarization.
\newblock In \emph{Proceedings of the 44th International ACM SIGIR Conference
  on Research and Development in Information Retrieval}, pages 1462--1471.

\bibitem[{Zhu et~al.(2021{\natexlab{b}})Zhu, Guo, Tian, Mao, and
  Han}]{zhu2021sumdocs}
Qi~Zhu, Fang Guo, Jingjing Tian, Yuning Mao, and Jiawei Han.
  2021{\natexlab{b}}.
\newblock Sumdocs: Surrounding-aware unsupervised multi-document summarization.
\newblock In \emph{Proceedings of the 2021 SIAM International Conference on
  Data Mining (SDM)}, pages 477--485. SIAM.

\end{thebibliography}
\bibliographystyle{acl_natbib}

\clearpage
\appendix

\section{Data Examples in \ourdata}
\label{sec_human_eval_case}
In Tables~\ref{table_example_ratings} and \ref{table_example_ratings2}, we show four data examples in \ourdata corresponding to different ratings in the human evaluation. While some of the examples are still of low quality after quality control, most of the filtered citation texts can serve as high-quality summaries.

\begin{table*}[t]
        \resizebox{2.05\columnwidth}{!}{
        \begin{tabular}{p{15cm}}
            \toprule
                \textbf{<Rating 1>} \\
                \textbf{Paper Title}: Congested traffic states in empirical observations and microscopic simulations \\
                \textbf{Paper Abstract}: We present data from several German freeways showing different kinds of congested traffic forming near road inhomogeneities, specifically lane closings, intersections, or uphill gradients. The states are localized or extended, homogeneous or oscillating. Combined states are observed as well, like the coexistence of moving localized clusters and clusters pinned at road inhomogeneities, or regions of oscillating congested traffic upstream of nearly homogeneous congested traffic. The experimental findings are consistent with a recently proposed theoretical phase diagram for traffic near on-ramps [D. Helbing, A. Hennecke, and M. Treiber, Phys. Rev. Lett. 82, 4360 (1999)]. We simulate these situations with a novel continuous microscopic single-lane model, the "intelligent driver model" (IDM), using the empirical boundary conditions. All observations, including the coexistence of states, are qualitatively reproduced by describing inhomogeneities with local variations of one model parameter. We show that the results of the microscopic model can be understood by formulating the theoretical phase diagram for bottlenecks in a more general way. In particular, a local drop of the road capacity induced by parameter variations has practically the same effect as an on-ramp. \\
                \textbf{Citation Text}: In a first approach, we use the well-known "intelligent driver model" (IDM) REF to show that the method works. \\
            \midrule
            
                \textbf{<Rating 2>} \\
                \textbf{Paper Title}: Probabilistic Model-Agnostic Meta-Learning \\
                \textbf{Paper Abstract}: Meta-learning for few-shot learning entails acquiring a prior over previous tasks and experiences, such that new tasks be learned from small amounts of data. However, a critical challenge in few-shot learning is task ambiguity: even when a powerful prior can be meta-learned from a large number of prior tasks, a small dataset for a new task can simply be too ambiguous to acquire a single model (e.g., a classifier) for that task that is accurate. In this paper, we propose a probabilistic meta-learning algorithm that can sample models for a new task from a model distribution. Our approach extends model-agnostic meta-learning, which adapts to new tasks via gradient descent, to incorporate a parameter distribution that is trained via a variational lower bound. At meta-test time, our algorithm adapts via a simple procedure that injects noise into gradient descent, and at meta-training time, the model is trained such that this stochastic adaptation procedure produces samples from the approximate model posterior. Our experimental results show that our method can sample plausible classifiers and regressors in ambiguous few-shot learning problems. \\
                \textbf{Citation Text}: They extended their approach by incorporating a probabilistic component such that for a new task, the model is sampled from a distribution of models to guarantee a higher model diversification for ambiguous tasks REF. \\

            \bottomrule
        \end{tabular}
        }
    \upv
    \caption{Examples in \ourdata with different quality ratings. }
    \label{table_example_ratings}
    \downv
    \end{table*}
    
\begin{table*}[t]
        \resizebox{2.05\columnwidth}{!}{
        \begin{tabular}{p{15cm}}
            \toprule

                            \textbf{<Rating 3>} \\
                \textbf{Paper Title}: A Generic Multi-Projection-Center Model and Calibration Method for Light Field Cameras\\
                \textbf{Paper Abstract}: Light field cameras can capture both spatial and angular information of light rays, enabling 3D reconstruction by a single exposure. The geometry of 3D reconstruction is affected by intrinsic parameters of a light field camera significantly. In the paper, we propose a multi-projection-center (MPC) model with 6 intrinsic parameters to characterize light field cameras based on traditional twoparallel-plane (TPP) representation. The MPC model can generally parameterize light field in different imaging formations, including conventional and focused light field cameras. By the constraints of 4D ray and 3D geometry, a 3D projective transformation is deduced to describe the relationship between geometric structure and the MPC coordinates. Based on the MPC model and projective transformation, we propose a calibration algorithm to verify our light field camera model. Our calibration method includes a close-form solution and a non-linear optimization by minimizing re-projection errors. Experimental results on both simulated and real scene data have verified the performance of our algorithm. \\
                \textbf{Citation Text}: Zhang et al REF proposed a multi-projection-center (MPC) model with six intrinsic parameters to characterize both conventional and focused LF cameras. \\
            \midrule
            
                            \textbf{<Rating 4>} \\
                \textbf{Paper Title}: Advancing Research Infrastructure Using OpenStack\\
                \textbf{Paper Abstract}: Abstract-Cloud computing, which evolved from grid computing, virtualisation and automation, has a potential to deliver a variety of services to the end user via the Internet. Using the Web to deliver Infrastructure, Software and Platform as a Service (SaaS/PaaS) has benefits of reducing the cost of investment in internal resources of an organisation. It also provides greater flexibility and scalability in the utilisation of the resources. There are different cloud deployment models -public, private, community and hybrid clouds. This paper presents the results of research and development work in deploying a private cloud using OpenStack at the University of Huddersfield, UK, integrated into the University campus Grid QGG. The aim of our research is to use a private cloud to improve the High Performance Computing (HPC) research infrastructure. This will lead to a flexible and scalable resource for research, teaching and assessment. As a result of our work we have deployed private QGG-cloud and devised a decision matrix and mechanisms required to expand HPC clusters into the cloud maximising the resource utilisation efficiency of the cloud. As part of teaching and assessment of computing courses an Automated Formative Assessment (AFA) system was implemented in the QGG-Cloud. The system utilises the cloud's flexibility and scalability to assign and reconfigure required resources for different tasks in the AFA. Furthermore, the throughput characteristics of assessment workflows were investigated and analysed so that the requirements for cloud-based provisioning can be adequately made. \\
                \textbf{Citation Text}: In REF , the authors focus on the use of a private cloud environment in order to improve the High Performance Computing (HPC) research infrastructure. \\
 
            \bottomrule
        \end{tabular}
        }
    \upv
    \caption{Examples in \ourdata with different quality ratings. }
    \label{table_example_ratings2}
    \downv
    \end{table*}

\section{Implementation Details}
\label{sec_implement}
Official results of the baselines are taken from prior studies when possible.
Model checkpoint selection is done on the validation set for every task.
The special token ``REF'' (used to indicate citation span) is removed from model output for all transfer experiments.
Paper abstract is used as input for all compared methods on SciTLDR.
We experimented with other prompts such as ``We'' and ``In REF'' but found ``This paper'' works the best.
FP16 is used in most experiments for training efficiency except for pre-training PEGASUS on \ourdata, with which it failed to learn. We use a batch size of 8. Hyperparameters like min/max generation length are generally set following prior work \cite{zhang2020pegasus}.

All the experiments are conducted with 1 Nvidia RTX A6000 GPU.
Pre-training on \ourdata only takes about 6.5h for BART and 10h for PEGASUS. 
The transfer experiments typically take less than 1h (time mostly spent on evaluation) as we use very few labeled data for training.
The codebase is based on Huggingface transformers \cite{wolf-etal-2020-transformers}.

\end{document}